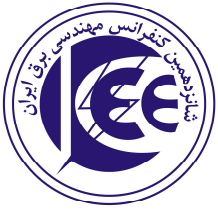

# Sparse Component Analysis (SCA) in Random-valued and Salt and Pepper Noise Removal

**Hadi. Zayyani, Seyyedmajid. Valliollahzadeh**
Sharif University of Technology
zayyani2000@yahoo.com, valliollahzadeh@yahoo.com

**Massoud. Babaie-Zadeh**
Sharif University of Technology
mbzadeh@yahoo.com

**Abstract:** *In this paper, we propose a new method for impulse noise removal from images. It uses the sparsity of images in the Discrete Cosine Transform (DCT) domain. The zeros in this domain give us the exact mathematical equation to reconstruct the pixels that are corrupted by random-value impulse noises. The proposed method can also detect and correct the corrupted pixels. Moreover, in a simpler case that salt and pepper noise is the brightest and darkest pixels in the image, we propose a simpler version of our method. In addition to the proposed method, we suggest a combination of the traditional median filter method with our method to yield better results when the percentage of the corrupted samples is high.*

## 1. Introduction

Impulse noise is caused by malfunctioning pixels in camera sensors, faulty memory locations in hardware or transmission in a noisy channel. The salt and pepper noise and the random valued-noise are the two common types of impulsive noises. In the salt and pepper noise, the salt noise is assumed to have the brightest gray level and the pepper noise has the darkest value of the gray level in the image. This assumption can help us to know the corrupted pixels in the images. In these cases the only hard task is to recover the original pixel of the image. But, in the general case of random-valued impulse noise, there is not any pre-assumption about the value of the impulsive noise. Therefore, the image denoising task in these cases is to detect the corrupted pixels and then correct them by the original pixel of the image. So, image denoising for random-valued impulse noises is more difficult than fixed salt and pepper image denoising. In this paper, we focus on the random value impulsive noise. However, we also present a version of our method in the case of salt and pepper noise.

The median filter is the most popular nonlinear filter for removing impulse noise [1]. However, when the noise level is high or when the random noise is available, some details and edges are smeared by the filter and the performance of the median filter decreases. Different remedies of the median filter have been proposed so far. They are the adaptive median filter [2], the median filter based on homogeneity [3], centre-weighted median filters [4] a generally family called decision-based methods. The so-called "decision-based" methods first identify possible noisy pixels and then replace them by using the median filter or its variants, while leaving all other pixels unchanged. Some of these two-stage methods deal with salt and pepper noise [5] and the others with the case of random-valued impulse noises [6].

In this paper, we do not separate the detection and correction steps similar to "decision-based" methods mentioned earlier. We use the compressibility of the images in the DCT domain which is used for image compression in JPEG standard. This compressibility gives us the necessary equation to exactly recover the impulsive noises or errors. Therefore, we use the transformed image to recover the noisy pixels. To recover the noisy pixels (or finding errors), we encounter an Underdetermined System of Linear Equations (USLE) whose sparse solution is to be found. This USLE problem can be solved by means of Sparse Component Analysis (SCA) methods [7]. In the SCA context, *m* sparse sources (which the most of their samples are nearly zero) and *n* linear observations of them are available. The goal is to find these sparse sources from the observations. The relation between the sources and the observations are:

**x = As** (1)

where **x** is the $n \times 1$ observation vector and **s** is the $m \times 1$ source vector and **A** is the $n \times m$ mixing matrix. *m* is the number of sources and *n* is the number of observations. In SCA, it is assumed that the number of sources is greater than the number of observations ($m > n$). So, the number of unknowns is larger than the equations. Therefore, this Underdetermined Linear System of Equations (ULSE) has infinite number of solutions. Fortunately, under conditions stated in [10], the sparsest solution of this problem is unique. This condition is that the number of active sources (non zero source) should be less than half of the number of observations ($\|s\|_0 < 0.5n$). By this assumption, the sparsest solution is unique and different algorithms to find this solution have been already proposed, including Basis-Pursuit (BP) [9], FOCUSS [10], smoothed-$l^0$ [11] and EM-MAP method [12]. The aim of this paper is to use the SCA methods in application of noise removal, especially for salt and paper noise and random-valued noise. The organization of the paper is as follows. Firstly, our SCA method is introduced in section 2, then this method in combination with popular median filtering is studied in section 3, and at last the simulation results will be discussed.

## 2. The proposed SCA method

### 2.1 Basic Idea

The basic idea is that, the representation of the image in the DCT domain is sparse because the most of the coefficients in the DCT domain are zero or near zero. We assume the noisy model as:

$\mathbf{X}_{N \times N} = \mathbf{S}_{N \times N} + \mathbf{E}_{N \times N}$ (2)

where $\mathbf{S}_{N \times N}$ is the original image and $\mathbf{E}_{N \times N}$ is the impulsive noise and $\mathbf{X}_{N \times N}$ is the noisy image (sub image). If we apply the DCT transform to both sides of equation (2), we have:

$T(\mathbf{X}) = T(\mathbf{S}) + T(\mathbf{E})$ (3)

where *T* is the DCT transform and has the following form:

$T(\mathbf{S}) = \mathbf{TST}'$ (4)

where *T* is the DCT transform matrix as defined below [1]:

$t(x,y) = \alpha(x)\cos((2y+1)\frac{x\pi}{2N})$

$\alpha(x) = \begin{cases} \sqrt{\frac{1}{N}} & x = 0 \\ \sqrt{\frac{2}{N}} & x \neq 0 \end{cases}$ (5)

We know that the block of $T(\mathbf{S})$ have many almost zero coefficients. To order this two dimensional matrix to a one dimensional vector with zeros at the end of the vector, we define the zigzag transform. This transform changes a two dimensional matrix to a one dimensional vector, similar to the JPEG standard. We assume that the coefficients of $Z(T(\mathbf{X}))$ are zero from $n+1$ to $m$. In this case *m* is the number of pixels in a sub image of size *N* and so is equal to $m = N^2$. Moreover, *n* is determined with the compression ratio of the sub image. If the compression ratio of the sub image is defined as $CR$, then the value of *n* is equal to $n = \frac{m}{CR}$. The general idea is to use this zeros to find the impulse noises (or errors). At first, we present the general case where the degraded pixels have random values and then switch to a simpler case where the salt and pepper assumption of noise are available.

### 2.2 Random value impulsive noise

By defining $\tilde{\mathbf{X}} \triangleq Z(T(\mathbf{X}))|_{n+1:m}$, and the previous assumption that transformed original image in the DCT domain is sparse, i.e. $Z(T(\mathbf{S}))|_{n+1:m} = 0$, we will have the following reconstruction formula to find the impulsive noises (or errors):

$\tilde{\mathbf{X}} = Z(T(\mathbf{E}))|_{n+1:m}$ (6)

If we are able to write the right hand of equation (6) in the linear form of $Z(T(\mathbf{E}))|_{n+1:m} = \mathbf{H}Z(\mathbf{E})$, then the problem of finding errors, converts to a classical SCA formulation as:

$$\tilde{X} = HZ(E) \quad (7)$$

Solving this SCA problem leads to zigzag transform of the errors. Taking the inverse zigzag transform yields the error image (both its value and its position). After subtracting the error image from the noisy image, the estimation of the original image is obtained. We call this method as "SCA method". The block diagram of this method is depicted in Fig. 1.

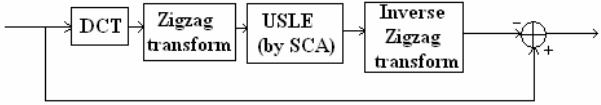

Fig. 1 The block diagram of our method

At first, we should find the matrix $H$ in terms of the DCT transform. To compute the matrix $H$, we use the 2-D transform equation in the general form [1]:

$$T(E)|_{(u,v)} = \sum_{x=0}^{N-1} \sum_{y=0}^{N-1} E(x,y) t(x,y,u,v) \quad (8)$$

Note that the $i$'th element of the $Z(T(E))$ equal to:

$$Z(T(E))|_i = T(E)|_{(u(i),v(i))} \quad (9)$$

where we can imagine the $[u(i),v(i)]$ as the inverse zigzag transform of the $i$'th 1-D element. From equations (8) and (9), we can write:

$$Z(T(E))|_i =$$
$$[t(0,0,u(i),v(i)), t(0,1,u(i),v(i)), .., t(N,N,u(i),v(i))]Z(E)$$
$$(10)$$

Therefore, $Z(T(E))$ can be written as $GZ(E)$, where the matrix $G$ is:

$$G_{ij} = t(u(j),v(j),u(i),v(i)) \quad (11)$$

From equations (6), (7), (9) and the preceding discussion, the matrix $H$ is $H = G(n+1:m, 1:m)$ where we use MATLAB matrix notation. The matrix $G$ is obtained simply from equation (11) and knowing that the DCT transform is separable of the form $t(x,y,u,v) = t(x,u)t(y,v)$. So, we have:

$$G_{ij} = t(u(j),u(i)).t(v(j),v(i)) \quad (12)$$

where $t(u(j),v(j),u(i),v(i))$ is defined in equation (5). Finally, the SCA problem in equation (7) can be solved by means of any SCA method such as MP, BP (or known as Linear Programming), smoothed-$l^0$ or EM-MAP. Since we should divide the image into the sub images and then solve the correspondence SCA problem with different $\tilde{X}$ and $H$, so a fast method for SCA is a necessity. Among the various methods, BP (or equivalently LP) and EM-MAP is rather complicated. Moreover, the MP method does not yield the accurate sparse solution of a SCA problem. However, a recently developed method called smoothed-$l^0$ [11] has the ability to provide a very fast and accurate estimation of the sparse solution. So, in our simulations we use this method.

### 2.3 Salt and pepper impulsive noise

In the salt and pepper impulsive noise, it is usually assumed that the salt noise is the maximum gray level (255) and the pepper noise is the minimum gray level (0) [5]. So, the places of noisy pixels are easily found by a simple comparison to these values (assuming that our image has not pixels with gray level 0 and 255). In [5], an adaptive median filter is used to detect the noisy pixels. But, in our paper, we assume that our image does not have pixels with gray level 0 and 255, and the noisy pixels are known by a simple comparison with the upper and lower gray levels. So, the only problem is to recover the original gray level of noisy pixels. Therefore, we propose a simpler version of our method. In this case we start from equation (7). Since the positions of errors are known, we can omit the columns of the matrix $H$ which we know that there is not any error at those places. So, equation (7) converts to the following formula:

$$\tilde{X} = H_{truncated} Z(E)_{nonzero} \quad (13)$$

After solving the above equation which is equal to solving a linear system of equations, the nonzero errors are obtained. In this case, the number of errors must be less than the size of the $\tilde{X}$ vector which is equal to $m-n$. The solution in these cases can be obtained via pseudo-inverse (where the unknowns are smaller than equations). We call this method the "Salt-Pepper SCA method" (SP-SCA).

## 3. The combined median-SCA-median method

Because of the good properties of the nonlinear filtering and especially median filtering in the image denoising applications, we suggest to use a combination of the traditional median filtering with our SCA method. When the noise level is low, the noisy pixels in a subimage are small and the median value of the sub image is not noisy. But, when the noise level is high the median value itself is a noisy pixel. So, the performance of the median filter is decreased. The median filter can be regarded as a pre-process to reduce the effect of the impulsive noise. After that, we can apply our SCA method. Moreover, in high level noise, this combination also cannot omit all the impulsive noises. Another median filter after our SCA method can omit the remaining impulse noises. So, the block diagram of this combination method is shown in Fig. 2.

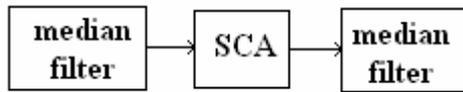

Fig. 2 The block diagram of combination of methods

## 4. Experiments

Three experiments were done to investigate our SCA method in image denoising when impulsive noise is present. In all experiments, the performance of our SCA method is compared with the median filter and also with the combination of methods. In the first experiment, we use the "SCA method" introduced in Sec. 2.2, and in the second and third experiments, we use the "salt-pepper SCA method" introduced previously in Sec. 2.3.

Our performance measure is the Peak-Signal-to-Noise Ratio (PSNR), defined as:

$$PSNR = 10\log_{10}(\frac{255^2}{\frac{1}{MN}\sum_{i,j}(s_{ij}-\hat{s}_{ij})^2}) \qquad (14)$$

### 4.1 Random-valued impulsive noise

In this experiment, random valued impulsive noise with different levels is added to the image. The results of the simulations are shown in Fig. 3. As we can see the combination of the methods has the best result in high level of noise (30% to 60% noise level). In addition to objective measures, the reconstructed images have good results up to 50% impulsive noise. Fig. 4 shows the corrupted image when 50% of pixels are corrupted with random-valued noise. Fig. 5 shows the reconstructed image.

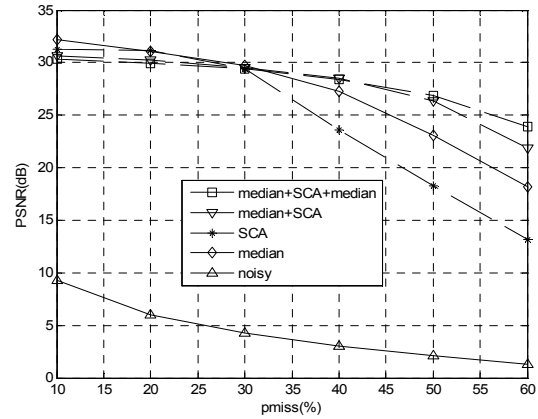

Fig. 3 The results for the random-valued noise

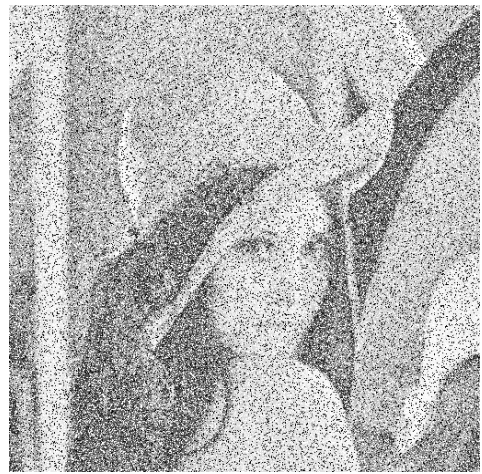

Fig. 4 The 50% random-valued noisy image

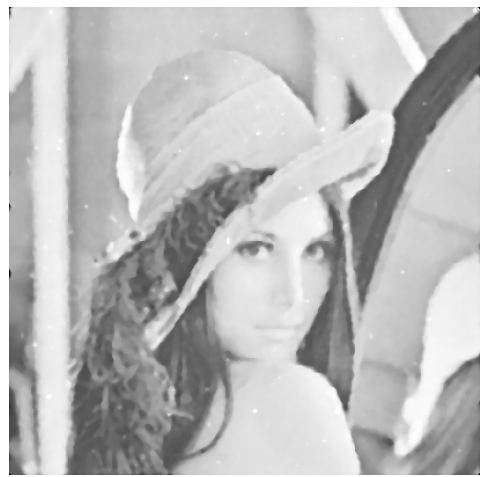

Fig. 5 The reconstructed image from 50% random-valued noisy image

### 4.2 Fixed gray level salt and pepper noise

In this experiment, it is assumed that only fixed gray level salt and pepper noise has corrupted the image (0 for pepper and 255 for salt). In this case, the image is reconstructed by the "salt-pepper SCA method" as introduced in Sec. 2.3. The results of various methods are depicted in Fig. 6. As it can be seen, our combination of methods has slightly better results especially at high noise levels. In this case, we can reconstruct the images ever it is corrupted by 60% salt and pepper noise. The noisy image and the reconstructed image in this case are shown in Fig. 7 and Fig. 8 respectively.

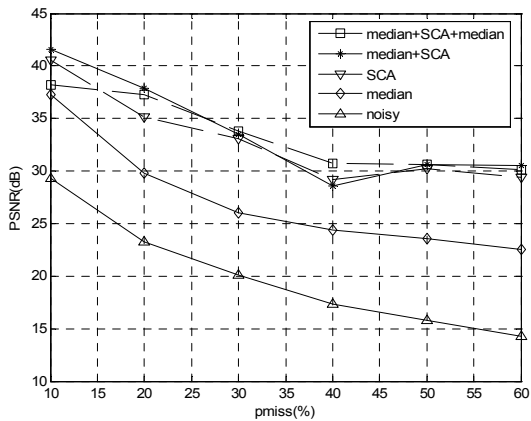

Fig. 6 The result for the fixed gray level salt and pepper noise

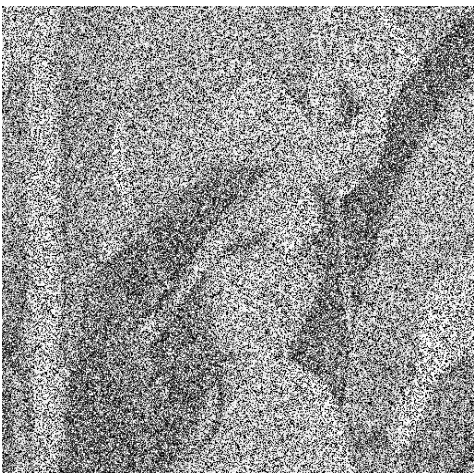

Fig. 7 The 60% fixed salt and pepper noisy image

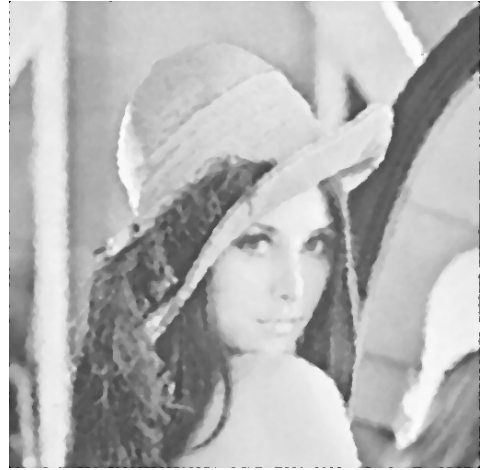

Fig. 8 The reconstructed image from 60% fixed salt and pepper noise

### 4.3 Missing sample

In this experiment, we assume that some pixels of the image are missed. So, those pixels are dark and have zero gray level. Similar to the previous experiment, the reconstruction of image is done by the "salt-pepper SCA method" as introduced in Sec. 2.3. The result of the simulations is shown in Fig. 9. In this case, the reconstruction was done appropriately up to 40% of missed samples. The missed-sample image and reconstructed image are shown in Fig. 10 and Fig. 11 respectively.

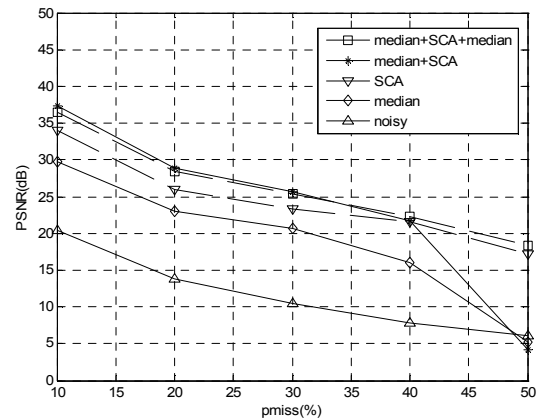

Fig. 9 The result for the missing sample experiment

### 5. Conclusion

In this paper, a novel method is proposed to remove impulsive noise from images. This method is essentially based on the sparsity of the images in the DCT domain. Using the nearly zeros in the DCT domain, an exact equation is provided to recover the impulse noises (or errors). To solve this equation, the smoothed-$l^0$ method [11] is utilized. In addition, in the simple case of fixed gray level salt and pepper noise, we present a new version of our method. To obtain better results when high level of noise is present, a combination of our SCA method with traditional median filtering is suggested. The simulation results show the efficiency of our method in the three cases of impulsive noise (random-value, fixed salt and pepper and missing sample).

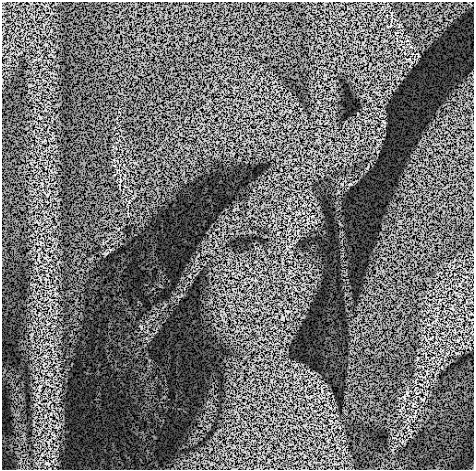

Fig. 10 The 40% missed-sample image

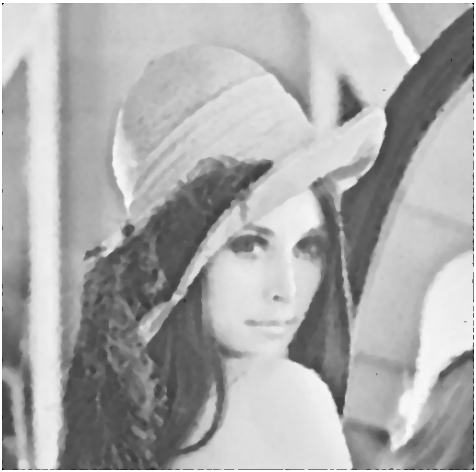

Fig. 11 The reconstructed image from 40% missed sample image

**Acknowledgement**

The authors would thank Advanced Communication Research Institute (ACRI) and Iran National Science Foundation (INSF) for financially supporting this work.